%% file: main.tex
\documentclass{article}

\usepackage[preprint]{neurips_2021}

\usepackage[utf8]{inputenc} 
\usepackage[T1]{fontenc}    
\usepackage{url}            
\usepackage{booktabs}       
\usepackage{amsfonts}       
\usepackage{nicefrac}       
\usepackage{microtype}      
\usepackage{xcolor}
\input{rnd_macros}
\input{automl_macros}

\usepackage{float}
\usepackage{wrapfig}
\restylefloat{figure}

\usepackage{xr-hyper}
\usepackage{hyperref}
\usepackage[capitalise,noabbrev]{cleveref}
\usepackage{subcaption}
\usepackage[export]{adjustbox}
\usepackage{gensymb}

\makeatletter
\newcommand{\printfnsymbol}[1]{%
  \textsuperscript{\@fnsymbol{#1}}%
}
\makeatother

\author{%
  Frederik Schubert\thanks{Equal Contribution} \\
  Leibniz University Hannover\\
  \texttt{schubert@tnt.uni-hannover.de} \\
  \And
  Theresa Eimer\printfnsymbol{1} \\
  Leibniz University Hannover\\
  \texttt{eimer@tnt.uni-hannover.de} \\
  \AND
  Bodo Rosenhahn \\
  Leibniz University Hannover\\
  \texttt{rosenhahn@tnt.uni-hannover.de} \\
  \And
  Marius Lindauer \\
  Leibniz University Hannover\\
  \texttt{lindauer@tnt.uni-hannover.de}
}

\makeatletter
\newcommand*{\addFileDependency}[1]{
  \typeout{(#1)}
  \@addtofilelist{#1}
  \IfFileExists{#1}{}{\typeout{No file #1.}}
}
\makeatother

\newcommand*{\myexternaldocument}[1]{
    \externaldocument{#1}
    \addFileDependency{#1.tex}
    \addFileDependency{#1.aux}
}

\myexternaldocument{appendix}

\newcommand{\subf}[2]{%
  {\small\begin{tabular}[t]{@{}c@{}}
 #1\\#2
\end{tabular}}%
}

\title{Automatic Risk Adaptation in Distributional Reinforcement Learning}

\begin{document}
\maketitle

\begin{abstract}
The use of Reinforcement Learning (RL) agents in practical applications requires the consideration of suboptimal outcomes, depending on the familiarity of the agent with its environment.
This is especially important in safety-critical environments, where errors can lead to high costs or damage.
In distributional RL, the risk-sensitivity can be controlled via different distortion measures of the estimated return distribution.
However, these distortion functions require an estimate of the risk level, which is difficult to obtain and depends on the current state.
In this work, we demonstrate the suboptimality of a static risk level estimation and propose a method to dynamically select risk levels at each environment step.
Our method \ours~(Automatic Risk Adaptation) estimates the appropriate risk level in both known and unknown environments using a Random Network Distillation error. 
We show reduced failure rates by up to a factor of $7$ and improved generalization performance by up to $14\%$ compared to both risk-aware and risk-agnostic agents in several locomotion environments.
\end{abstract} 

\section{Introduction}
Many real-world applications of Reinforcement Learning (RL) are risk sensitive, including robotics, autonomous driving and healthcare.
This means that some states such as a car crash are associated with a very high cost and must be avoided not only during deployment but already while training an RL agent.
Furthermore, the risk of failure might change because of variations in the agent's task~\citep{pinto-icml17,zhao-ssci20}.
So in order to be applicable in practice, agents need to avoid fatal mistakes in particular in the face of a changing environment.

Prior work assumed that information about risky elements in an environment, such as failure states, is explicitly provided via a model of the environment \citep{garcia-jmlr15, turchetta-neurips20, jansen-concur20}.
However, knowledge about the target domain might not always be readily available. Therefore, there is a need for agents that can learn a risk-aware behaviour on their own. 
To reduce the need for external information, we propose that an agent can extract safety-relevant information from its own interactions with the environment.

One natural approach is to consider a risk-level of an action in a given state for making a decision.
The problem faced in practice is how to choose an appropriate risk-level without having a lot of prior knowledge about the environment. 
Too conservative behavior could lead to suboptimal exploration and therefore reduced reward. In contrast, too risky behavior could lead to many failure cases during training. 
While it is possible to use intuition in order to select such risk-levels, we demonstrate both the difficulty of selecting an appropriate risk level and the suboptimality of an one-size-fits-all approach to risk estimation. 

\begin{figure}[t]
    \centering
    \includegraphics[width=0.8\textwidth]{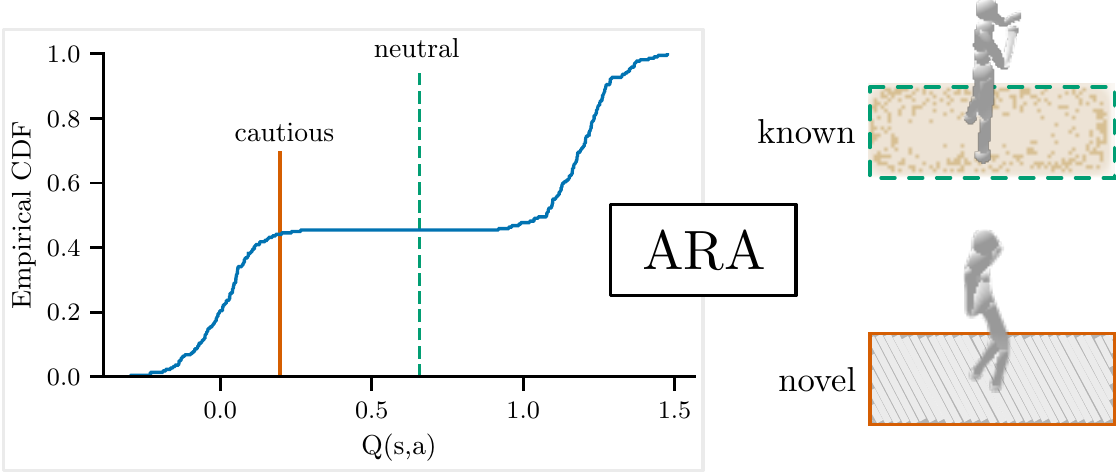}
    \caption{Illustration of an agent that is faced with changed dynamics during training and its effect on the value estimation through \ours{}. Shown is the estimated empirical CDF of Q-Values for an action $\action$ in state $\state$. There are two equally likely outcomes. However, the estimated average return (dashed green) is less likely with smaller ground friction. As the current state in the novel environment (solid brown) has not been seen before, \ours{} causes the agent to act more cautiously.}
    \label{fig:teaser}
\end{figure}

To relieve users from the burden of choosing a risk-level, we propose Automatic Risk Adaptation (\ours)\footnote{Code can be found in the supplementary and will be made public upon acceptance.}. 
It adjusts the agent's risk-aware behaviour under changing risk levels without additional information (see \cref{fig:teaser}).
The idea of ARA is based on the parametric uncertainty of the agent's predictions to adapt its risk-awareness, thus encouraging more cautious behaviour in states that have not been visited as often.
As an additional advantage over prior work~\citep{choi-corr21}, it does not add significant overhead to training as it requires no additional episode evaluations. 

Since risk-awareness is especially interesting for agents moving around in an open and potentially continuously changing environment, we focus on robot tasks to study the properties of \ours{}. In particular, we demonstrate the effectiveness of \ours{} on 
a variety of robot locomotion tasks.

Our \textbf{contributions} therefore are:
\begin{enumerate}
    \item We show exemplary that static risk levels are suboptimal for generalization to variations of an environment on a windy grid-world.
    \item We propose a risk sensitivity regulation mechanism, called \ours{}, to apply distributional RL algorithms in real world applications.
    \item \ours{} reduces failure rates in risk-sensitive tasks by up to $7$ times compared to static CVaR policies.
    \item Our method improves generalization performance by up to $14\%$ across varying risk levels.
\end{enumerate}

\section{Related Work}
Safety in risk-sensitive settings is a well recognized challenge in Reinforcement Learning~\citep{garcia-jmlr15}. 
One way to address this challenge is the risk-aware distortion of quantile functions of the return distribution~\citep{dabney-icml18}, as for example shown by \cite{ma20} and \cite{keramati-aaai20}. 
Such approaches often use Conditional Value at Risk (CVaR)~\citep{rockafellarOptimizationConditionalValueatrisk2000} as it is an effective method to optimize worst-case performance.
However, these works keep the risk level $\alpha$ fixed which inhibits generalization to different risk levels.
In contrast, \cite{lyu-aamas20} use a simple schedule to adapt the risk level over time in a multi-agent setting in order to foster cooperation.
While this is beneficial for overall performance, a hand-crafted schedule, similarly to a hand-crafted static value, can only ever be an approximation of the true risk and is therefore constrained in the domains it can be applied to.
We aim to improve over both by providing a dynamic risk-adjustment method that is sensitive to change in both the policy and environment and thus broadly applicable.
\cite{choi-corr21} employ a dynamic approach by conditioning the policy on the risk level $\alpha$. 
Their method requires policy evaluation and training across different uniformly distributed $\alpha$ levels, which dramatically increases the number of observations and therefore the computational complexity needed for training.\footnote{We were not able to compare against \cite{choi-corr21} as their code is not publicly available.} 
In contrast, our method requires very little overhead and no additional episode evaluations for risk level estimation.

Where domain knowledge is available, it can be used to estimate the appropriate risk level.
In model-based RL, the world model can function both as a simulation environment and a source of knowledge about the risk of the setting \citep{jansen-concur20, yu-fr21}.
%
Input from a teacher is another way to convey risk in RL.
In active learning, agents can query human input more frequently in risky parts of the state space \citep{brown-corr19}.
Similarly, a teacher can be used to teach the agent environment constraints like fatal actions without the agent needing to experience them \citep{turchetta-neurips20}.
We assume that such domain knowledge or teacher experience is not generally available and thus aim to find other sources for risk-adaption.

As more conservative action choices conflict with exploring the environment, safe exploration is a field of research that complements risk-aware policies, such as \ours{}, especially in environments that require thorough exploration of the state space.
Just as for risk-aware action selection, there are many possible approaches to solve this problem, from optimism in the face of uncertainty \citep{keramati-aaai20} to learnt exploration mechanisms using e.g. Gaussian Processes \citep{sui-icml15}.

\section{Risk-Aware Distributional Reinforcement Learning}

Before describing the specifics of \ours{}, we provide the necessary background on quantile function distortion for risk-awareness in distributional RL agents.

\paragraph{Distributional Reinforcement Learning (DistRL)}approximates not only the expected return with respect to a given state-action pair, but the whole return distribution \citep{bellemareDistributionalPerspectiveReinforcement}.
This enables DistRL agents to integrate the uncertainty of the return into their decision process.

Distributional Deep Q-Learning via Quantile Regression \citep{dabney2018distributional} learns to estimate the quantile function of Q-values for a given state-action pair. 
This is done by modeling the Q-value distribution in state $\state$ for an action $\action$ as a random variable $Z(s, a)$. 
Let $Z_{\tau}$ denote the inverse cumulative distribution function (CDF) of $Z$.
If we query $Z_{\tau}$ using $\tau \sim \mathcal{U}(0, 1)$, we obtain a Q-value sampled from $Z$.
Each percentile rank $\tau$ corresponds to a possible Q-value for a pair $(\state, \action)$ that is distributed according to $Z$, so in expectation:
\begin{equation}
    Q(\state, \action) = \mathbb{E}_{\tau\sim \mathcal{U}(0, 1)}[Z_{\tau}(\state, \action)]
    \label{eq:q_z}
\end{equation}
Estimating the quantiles of the Q-value distribution lifts the constraint of setting fixed Q-value ranges of previous approaches such as the C51 DistRL agent \citep{bellemareDistributionalPerspectiveReinforcement}.

While in \cref{eq:q_z} all possible Q-values in a state $\state$ are still equally weighted, it is also possible to distort the distribution using a non-linear distortion function $\beta: [0, 1] \rightarrow [0, 1]$ \citep{balbasPropertiesDistortionRisk2009}.
This results in a more pessimistic or more optimistic estimate of the Q-value distribution in a given state, depending on the chosen function.
\begin{equation}
    Q_{\beta}(\state, \action) = \mathbb{E}_{\tau}[Z_{\beta(\tau)}(\state, \action)]
    \label{eq:distr_q}
\end{equation}

Distorting the quantile distribution can thus control the kind of policy that a DistRL agent aims to learn, e.g. one that optimizes the worst-case performance in a given environment.
\paragraph{Quantile Function Distortion}~\citep{chow-neurips14} has frequently been used in distributional RL for risk-sensitive settings \citep{dabney-icml18,singh-l4dc20}. 
In contrast to other forms of risk assessment, e.g. worst outcome performance or uncertainty measures, which are often hard to measure or take a considerable amount of compute, distortion functions can be easily applied to distributional value functions.
A common example is CVaR~\citep{chow-neurips14}, also known as \emph{expected shortcoming}.

CVaR describes the expectation that the value of a random variable, in this setting the Q-value, is lower than the value at risk (VaR) at a given confidence level $\alpha$.
The VaR in turn is an originally economic risk measure that describes the maximum financial loss with probability $\alpha$.
We can view it as a measure of regret in this application.
The CVaR is used as a mapping $\beta$ to distort the quantile function for action selection and compute the gradients for the value function updates. 
\begin{equation}
    \beta_{\text{CVaR}}(\tau;\alpha) = \tau \cdot \alpha
    \label{eq:cvar_quantile_transform}
\end{equation}
Therefore we take an estimation of the regret with a given probability into account and make more conservative choices overall, as indicated by the confidence level $\alpha$. 
We can also view this confidence level as a risk-level since high confidence implies a low risk setting.

\section{ARA: Automatic Risk Adaptation}

While CVaR is an 
easy-to-use and effective risk measure, it has an additional hyperparameter in $\alpha$. 
As $\alpha$ is the risk level of the distortion function, it has a significant impact on the success of training and needs to be set correctly for the problem at hand. 
For this reason, we propose a way to automatically adapt the risk sensitivity of the setting depending on the agent's current situation, allowing for an adaptive behaviour that is cautious in risky areas of the state space and less so in safe zones.

\subsection{Static Risk Levels are Suboptimal for Generalization}
\label{subsec:static_risk_suboptimal}
\begin{figure}[h]
\centering
\begin{tabular}{c c}
\subf{\includegraphics[valign=T,width=25mm]{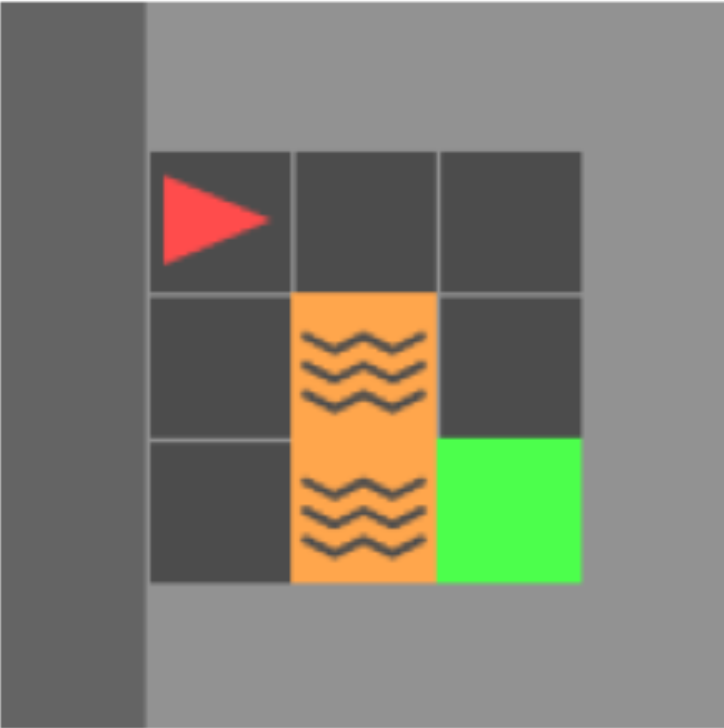}}{}
&
\begin{tabular}[t]{l c c c}
  \toprule
  Setting & Best Choice & Mean Failures & Worst Choice\\
  \midrule
  Light south wind  & 0.00 $(\alpha: 0.4)$ & 0.0 & 0.0\\
  \midrule
  Strong south wind & 0.04 $(\alpha: 0.5)$& 0.05 & 0.07 \\
  Light north wind & 0.22 $(\alpha: 0.1)$& 0.27 & 0.33 \\
  Light east wind & 0.34 $(\alpha: 0.9)$ & 0.44 & 0.49\\
  \bottomrule
  \label{tab:grid_dist}
\end{tabular}
\end{tabular}
\caption{\textbf{Left:} Windy GridWorld with Lava Crossing. \textbf{Right:} Differences in failure rates for several values of $\alpha$ over $100$ evaluations of the same value function under CVaR risk distortion with different training (first row) and test settings. The columns show the best, average and worst failure rate that a CVaR policy reaches.}
\label{fig:winds}
\end{figure}

To show that statically chosen risk levels in quantile function distortion are generally not a good choice in varying environment conditions,
we use the GridWorld LavaGap \citep{gym-minigrid} environment  with added wind that can turn the agent (see \cref{fig:winds}). 
The agent can turn left or right and move forward. 
There is no time limit to reach the goal state, so the only way for the agent to fail is by falling into the lava.

In this setting we can vary both wind direction and wind strength to adjust the risk level.
We compute the tabular optimal value function for light south wind that turns the agent with a likelihood of $25\%$ and construct a categorical distribution with $50$ values for $\tau$ as a distributional value function.
Additionally, we distort the value function with CVaR 
using values for $\alpha$ in increments of $0.1$ between $0$ and $1$.

We evaluate $100$ episodes for each policy in settings with greater wind strength ($90\%$ instead of $25\%$, 'Strong south wind') and different wind directions ($90\degree$ turn in direction with 'light east wind', $180\degree$ with 'light north wind'). 
We can see the difference the choice of risk level makes in \cref{fig:winds}.

We observe that different choices for $\alpha$, even in this very simple setting, lead to different failure rates, with a widening disparity between the mean performance and the best one.
While in the original 'light south wind' setting, there was no difference, it increases to around $2\%$ for stronger wind to $5\%$ and eventually $10\%$ with winds from different directions the agent has not trained on.
Furthermore, the mean performance of all choices for $\alpha$ moves closer to the performance of the worst choice with rising risk level.
This means the performance gap between different values for $\alpha$ is widening significantly in riskier settings with unfamiliar wind directions, making the choice more important.

The values that performed best in each setting do not seem to follow an obvious pattern. While the best failure rate in the original 'light south wind' variation was achieved with $\alpha=0.4$, it increased to $0.5$ with stronger wind and changed drastically to $0.1$ and $0.9$ respectively for east and north wind.

This shows that users will not be able to easily identify the correct value for $\alpha$ even in such a simple setting, limiting the pratical use of CVaR. 
At the same time our results emphasize that correctly choosing $\alpha$ is a major factor for performance and failure prevention, especially in high risk settings.

To effectively take advantage of CVaR policies across settings, $\alpha$ should therefore be dynamically adapted to the current environment conditions, like changes of wind direction or strength in the example above.
As this complicates the selection process even further, however, we present an approach to find and adjust $\alpha$ automatically on the fly.

\subsection{Dynamic Risk Level Estimation}

In order to correctly adapt $\alpha$ in any given risk-distorted DistRL policy, we propose using Random Network Distillation to approximate the environment's current risk level.

\paragraph{Random Network Distillation (RND)}~\citep{burda-iclr19} provides a measure of uncertainty that has previously been applied to the exploration problem in environments with sparse rewards.
It uses a randomly initialized frozen target network $f$ and a predictor network $g$.
Given the same input, the predictor aims to match the random features of the target network.
Because both networks receive the same input and the target network is in the same function class as the predictor, the prediction error $u$ on the training data decreases as the amount of samples increases:
\begin{equation}
    u(s) = \pnorm{f(s) - g(s)}^2
    \label{eq:rnd_parameter}
\end{equation}
However, changes in the training distribution manifest themselves in an increase in the prediction error.
This yields an uncertainty estimate for any state based on the current policy and environment dynamics.
As we expect to see failure states less often in training due to penalization in the reward function, transitions around those states will be seen less often and produce a higher uncertainty.

\begin{wrapfigure}{R}{0.5\linewidth}
    \centering
    \includegraphics[width=\linewidth]{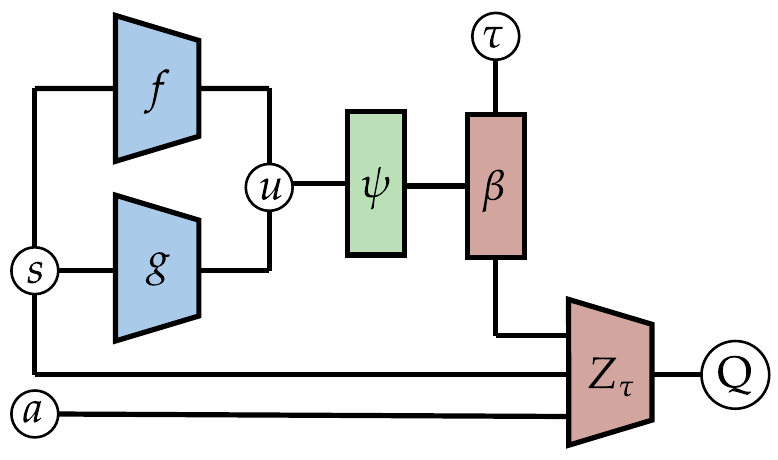}
    \caption{Components of \ours{}. The target $f$ and the predictor $g$ of the RND are shaded in blue, the uncertainty to risk mapping $\psi$ of \ours{} in green and the DistRL components in red.}
    \label{fig:flowchart}
\end{wrapfigure}

\paragraph{From Uncertainty to Risk-Level}

As discussed above, quantile function distortion is an effective way to integrate risk-awareness into any distributional RL agent predicting state-action values at the cost of manually tuning its risk level.
With \ours{}, however, we use RND to estimate the appropriate $\alpha$.
RND provides an estimate of how often a state has been visited in training.
If there are many unknown states in a given environment, we prefer the agent to act conservatively in order to avoid fail states.
Additionally, the agent is incentivized to avoid fail states during training because of the associated negative reward. 
Therefore the likelihood of a state with higher associated RND error being a fail state is comparatively large and we want the agent to increase its risk level in its proximity.

To reflect this in the DistRL algorithm, we use the RND error for the parameter $\alpha$ in the distortion function $\beta$ in \cref{eq:distr_q}:
\begin{equation}
    \beta_{\text{ARA}}(\tau) = 
    \beta_{\text{CVaR}}(\tau;\psi(u))
    \label{eq:beta_ara}
\end{equation}
with $u$ being the RND error and $\psi(u) = e^{-u}$. 
In order to keep $u$ on a consistent scale, we normalize it by its running average estimate.

The RND architecture is an additional hyperparameter of our algorithm, though
far less subject to changes in the current setting.
We have found the architecture as originally proposed by \cite{burda-iclr19} to work well.
We explore the influence of $\psi$ on the performance of our method in Appendix~\ref{app:rnd_mapping}.

Using RND for risk estimation in this way is independent of the action and state space size. 
Furthermore, the risk level estimation is appropriate for the policy even in a test setting as we are not introducing a second learnt element but base the estimation on the policy itself. 
Thus, \ours{} can provide a reliable risk estimation in many settings without significant training overhead.

\section{Experiments}

We show improved generalization performance and lower failure rates in high risk settings compared to conventional static risk level estimations using \ours{} in combination with DSAC~\citep{ma20} on several 3D locomotion tasks.

\subsection{Robustness to Changing Dynamics}
\label{subsec:learning_dynamics}

Given enough training data, we expect an RL agent to benefit from training on a diverse distribution of its task \citep{pmlr-v97-cobbe19a}.
When faced with a variation of the problem during training, however, risk-aware agents have to adjust their behaviour, i.e. their risk level $\alpha$, to take these different dynamics into account, as we showed on a small example in \cref{subsec:static_risk_suboptimal}.
This is of course crucial for real-world applications, where we usually cannot guarantee a perfectly stable training environment.
Here we demonstrate that this principle also applies to more complex problems and that the risk adaption through \ours{} therefore can accelerate training and robustness.

\begin{figure}[h]
    \centering
    \includegraphics[width=\linewidth]{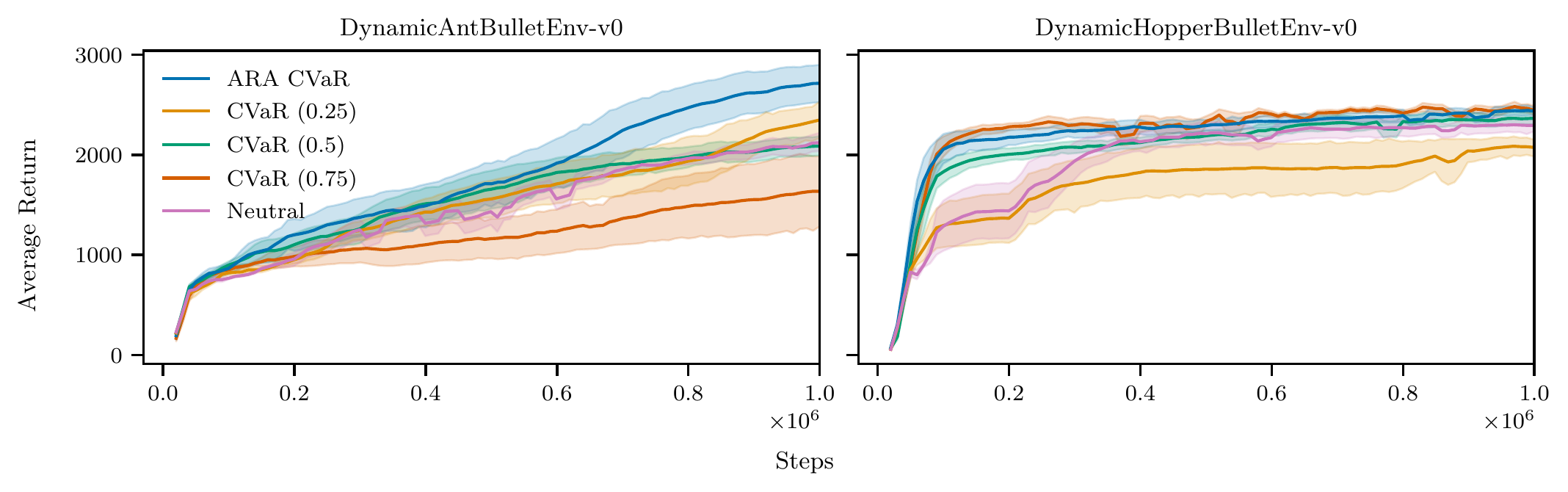}
\caption{Average return ($\pm$ standard error) over 50 episodes (evaluated at each $10\,000$ step interval) during training with changing dynamics. 
}
\label{fig:results_dynamic_walker}
\end{figure}

\paragraph{Experimental Setup}
We evaluate \ours{} on variations of the \emph{AntBulletEnv-v0} and \emph{HopperBulletEnv-v0} environments that are based on the \emph{PyBullet} physics engine \citep{pybullet} with changing environment dynamics.
We vary the friction parameters of the feet and the mass\footnote{The mass and friction parameters are called \texttt{mass}, \texttt{lateralFriction}, \texttt{spinningFriction} and \texttt{rollingFriction} in PyBullet and can be changed via the \texttt{changeDynamics} method.} of the torso randomly by up to $20\%$ of their default values in each episode.
For a more thorough exploration of different variation levels, see \cref{app:ablation}.
Since the environment dynamics are not part of the agent's observation, it needs to learn to distinguish between different levels of friction and mass purely through its interactions with the environment.
As defined by the environments, the reward is comprised of a movement cost, cost of collision and a reward for moving forward.

Falling down is an explicit failure case in these environments, but this is not the only way overly risky or cautious behaviour can lead to performance reduction.
It is just as important to judge movement speed and direction appropriately for the current friction and mass levels in order to move as quickly and safely possible.
Therefore, risk-awareness is closely connected to the reward in this setting.

We compare to a non-risk-aware DSAC agent and a DSAC agent with added static CVaR distortion with an $\alpha$ of $0.25$ (as is common in the field \citep{ma20,keramati-aaai20}), $0.5$ and $0.75$.
This allows us to judge \ours{} against agents of varying static risk levels.
All results show averages and standard error over $5$ runs with different random seeds.
For hardware specifications and algorithm hyperparameters, please refer to Appendix \ref{app:hypers}.

\paragraph{Empirical Results} are shown in \cref{fig:results_dynamic_walker}.
On \emph{DynamicAntBullet}, \ours{} is able to achieve a better final performance than both the statically risk-aware CVaR agents and the risk neutral agent by at least $14\%$.
Clearly a higher risk level is beneficial for this task, as an $\alpha$ of $0.25$ does improve final performance compared to the higher $0.5$ or the neutral agent slightly.
$\alpha$ of~$0.75$ on the other hand is not a good choice in this setting, as the agent here can only reach about $60\%$ of \ours{}'s performance. 

While \ours{}'s final performance in the \emph{DynamicHopperBullet} is only slightly better than neutral and CVaR with $\alpha$ of $0.5$ and around the same as an $\alpha$ of $0.75$, it reaches this performance as fast as the best performing static CVaR with $\alpha=0.75$. In addition, \ours{} is three times faster than the neutral agent.
In contrast to the \emph{DynamicAntBullet}, the more conservative CVaR with $0.25$ falls short on this task.

Overall we observe that the performance ranks of the static CVaR agents are reversed on these environments, with the \emph{DynamicAntBullet} requiring a more cautious approach compared to the \emph{DynamicHopperBullet}.
Even so, the risk-awareness is beneficial on both as demonstrated by the neutral agent performing only averagely.
We can conclude that the Hopper robot is either easier to control or less likely to fail, giving the tasks a different risk distribution.
This is reflected in the average risk parameter which is $10\%$ higher (i.e. less risky) at the end of training for \emph{DynamicHopperBullet}.

Our agents with static CVaR distortion can not outperform \ours{} on either environments or match its overall performance. 
As they lack \ours{}'s ability to detect the changing dynamics and automatically adapt the risk level, they are limited to performing well on only a subset of environments and environment conditions.
We can see this effect on CVaR with $\alpha$ of $0.25$ and $0.75$, that each performs significantly better in one environment than the other.
The fact that they also cannot adapt to the changing conditions furthers the gap between \ours{} and the best static value we tested, by not only matching the best static CVaR agent in each environment but beating it considerably in the case of the \emph{DynamicAntBullet} environment.
Overall, our approach is more robust and able to generalize over dynamic changes, regardless of the inherent risk level in the environment.

\paragraph{Further Insights} To confirm whether the gains of \ours{} stem from an improved estimation of the returns under the changing dynamics, we analyze the average Q-value estimation error of the algorithms during training.
For this, we plot the average difference between the maximum empirical discounted return and the agent's estimated Q-value.

\begin{figure}[h]
    \centering
    \includegraphics[width=\linewidth]{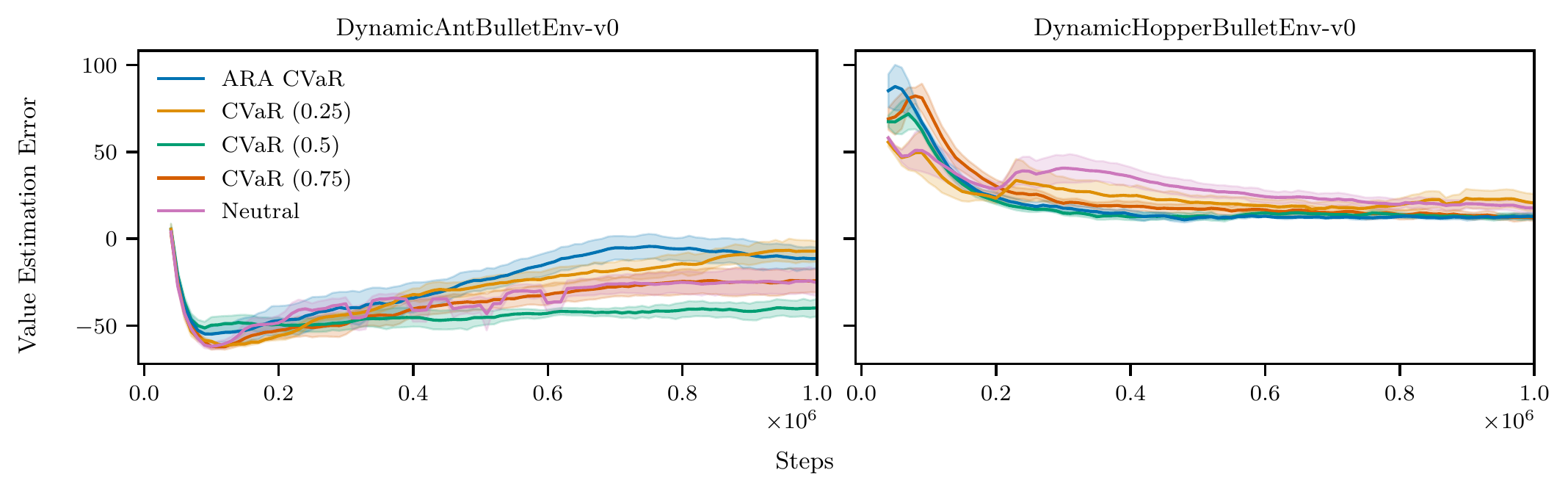}
    \caption{
    Average Q-value estimation error during training ($\pm$ standard error). Error close to $0$ is better.}
    \label{fig:value_estimation_error}
\end{figure}

From \cref{fig:value_estimation_error} it is apparent that \ours{} leads the RL agent to more accurately estimate the Q-values.
In the \emph{DynamicAntBullet}, \ours{} and the static CVaR with $\alpha=0.5$ show a less pronounces overestimation of the Q-values in the beginning than the other agents.
However, only \ours{} is able to further reduce the value estimation error later in training at around $600,000$ steps, resulting in a better performance.

Interestingly, the error in Q-value estimation does not directly correlate to the static CVaR agents' performance in this environment. 
The CVaR with $\alpha$ of $0.25$ should perform as well as \ours{} and $\alpha$ of $0.75$ should performs much better than $0.5$ according to how well they approximate the Q-values. 
We suspect this is due to the CVaR distortion preventing the policy from acting optimally in situations the agent is confident in whereas \ours{} will trust the policy in these states.

The results for the \emph{DynamicHopperBullet} are not as clear, but again, \ours{} and the CVaR with $\alpha$ of $0.5$ and $0.75$ agents have the smallest estimation errors.
It should be noted that the neutral DSAC has more variance in its Q-value estimate in both environments.

Our hypothesis for the low Q-value estimation error of \ours{} is that the agent explores a part of the state space until it has seen enough samples to act less cautiously and discover novel states.
So even if \ours{} does not directly affect the Q-networks through any gradient updates (as it is only used in the update of the policy network), it 
enables the agent to safely experience the environment.

\subsection{Effects of Adaptive Risk Levels during Training}
\label{subsec:eval_risk_sensitive}

Dynamic risk adaption has an additional benefit besides acting according to the current setting.
It is able to prioritize risky or cautious behaviour according to how much the agent has learnt already.
We want to be more cautious in the beginning of training, but trust the agent more as it performs better.

\paragraph{Experimental Setup} We show this effect on the \emph{BipedalWalkerHardcore-v3} locomotion environment \citep{gym}.
In this environment the agent has to learn to walk over a terrain that it observes via several sensors, the reward corresponding to the distance traveled (at most $300$).
The terrain varies between episodes, so the agent has to learn to generalize over possible obstacles on the ground.
If the agent falls, the episode is counted as a failure and the agent receives a reward of~$-100$.

Unlike in the scenarios in \cref{subsec:learning_dynamics}, reward and failure rate are not as closely connected here.
The reward only takes distance into account, so therefore failing very early on a few runs but finishing the rest may result in the same performance as failing on most runs near the end.

\begin{figure}[h]
\begin{subfigure}[b]{0.48\linewidth}
    \centering
    \includegraphics[width=\linewidth]{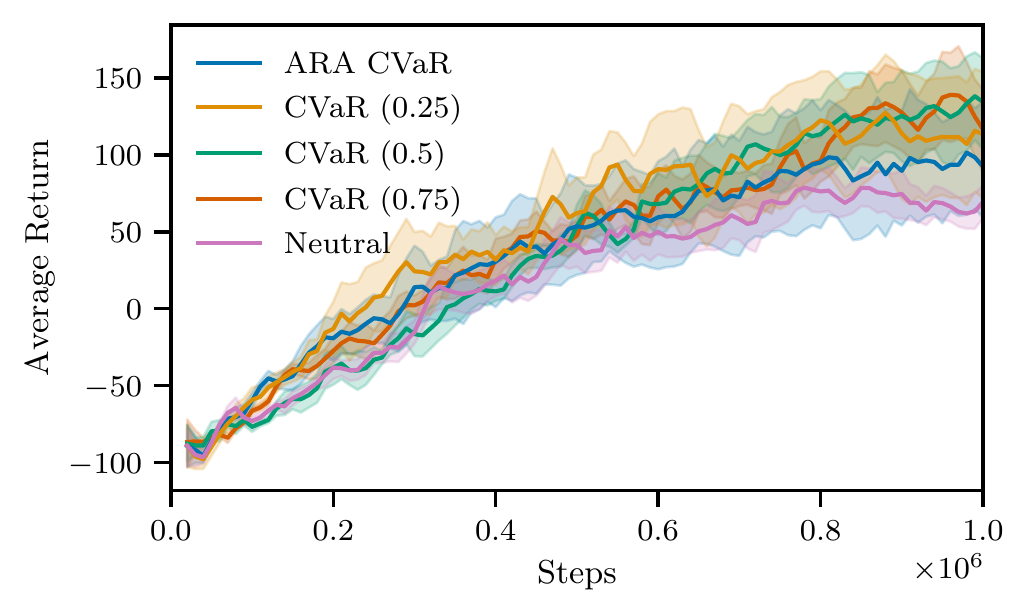}
\end{subfigure}
\begin{subfigure}[b]{0.48\linewidth}
    \centering
    \includegraphics[width=\linewidth]{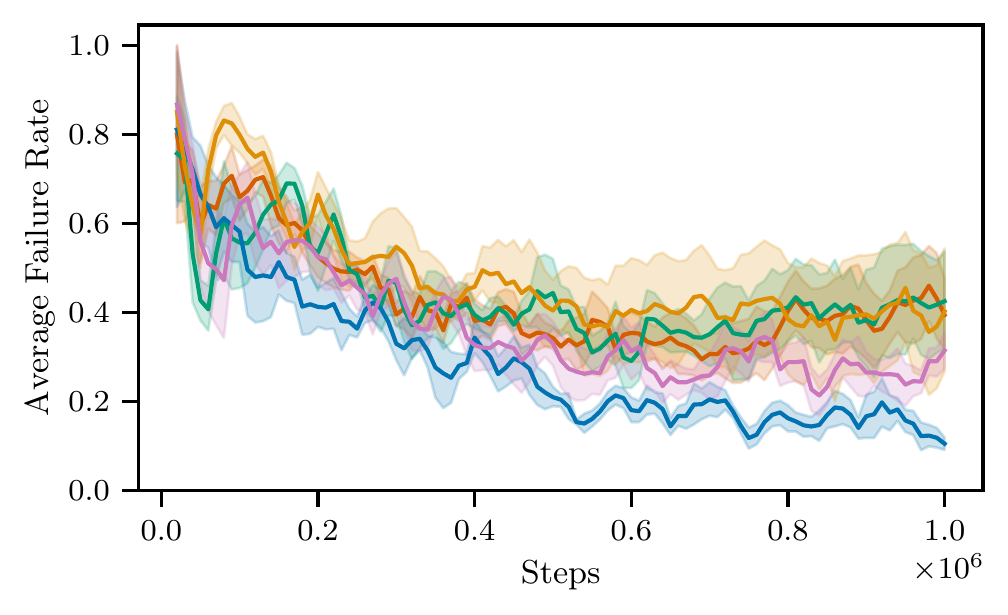}
\end{subfigure}
\caption{Average return (left) and failure rate (right) on \emph{BipedalWalkerHardcore-v3} over 50 episodes (evaluated at each  $10,000$ step interval) during training of Neutral DSAC, DSAC with fixed CVaR $\alpha$ ($0.25$, $0.5$ and $0.75$) and DSAC with \ours{}. The shaded area indicates the standard error over $5$ runs.
}
\label{fig:results_bipedalwalkerhardcore}
\end{figure}

\paragraph{Empirical Results} In \cref{fig:results_bipedalwalkerhardcore} the average return and failure rate of \ours{} are shown during training.
We see that there is indeed a trade-off between performance and safety in this scenario. 
It is apparent that a higher average return performance does not always lead to lower failure rates as here, all approaches are fairly close in terms of reward, with the CVaR using $\alpha$ of $0.5$ and $0.75$ outperforming \ours{}, and $\alpha=0.25$ and the neutral agent performing worst.

\begin{wrapfigure}[16]{R}{0.5\textwidth}
    \centering
    \includegraphics[width=\linewidth]{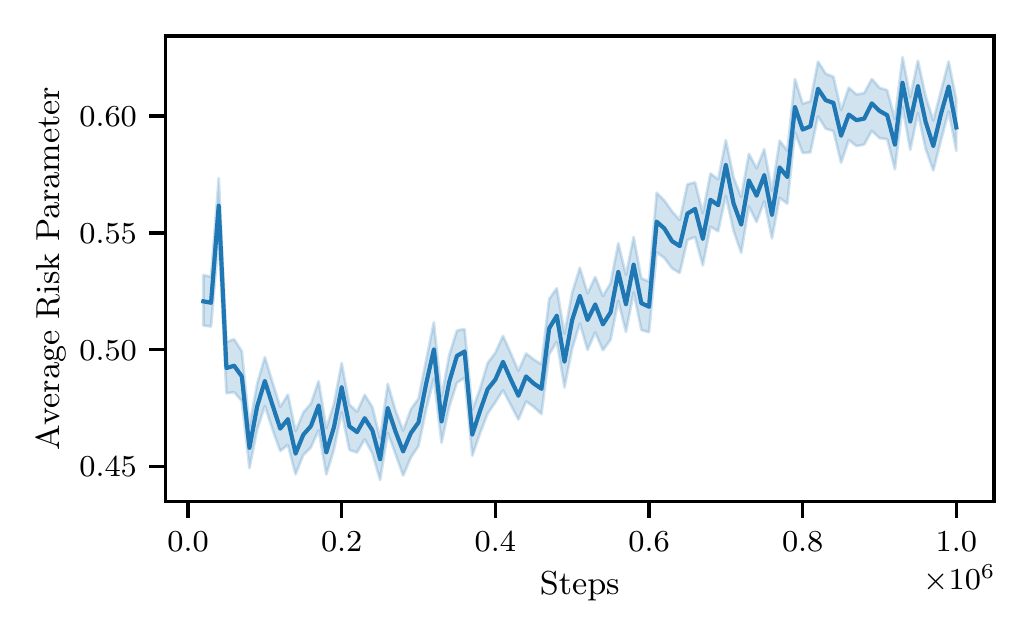}
    \caption{Average risk parameter of \ours{} over 5 seeds on \emph{BipedalWalkerHardcore-v3} ($\pm$ standard error).}
    \label{fig:bipedalwalkerhardcore_risk}
\end{wrapfigure}

\ours{} is able to adapt the risk level to the current state and achieves consistently lower failure rates than the baselines, however.
It has a lower failure rate than every other method at every point during training and eventually improving by a factor of $4$ and $7$ over the neutral and static risk-aware agents respectively.
As \ours{} uses the parametric uncertainty to adapt the risk level, it can learn to significantly reduce failures even in this setting where reward function and failure rate are not as closely connected.
 
The risk parameter ($\psi(u)$ in \cref{eq:beta_ara}) during training shows in \cref{fig:bipedalwalkerhardcore_risk} how risk-level and policy performance interact.
After a drop in the beginning that is caused by the robot reaching novel states, it slowly rises as the agent 
learns to control the robot.
The stagnation in the end is caused by new obstacles that the agent has to overcome to continue.

Overall, \ours{} successfully matches the risk parameter to the policy development, making the training much safer while obtaining similar performance to the other agents.
The adaptive risk parameter supports the safe training of the agent, regulating trust in the learnt policy according to its quality.

\newpage
\section{Limitations}
\label{limitations}

\begin{figure}[h]
    \centering
    \includegraphics[width=0.45\linewidth]{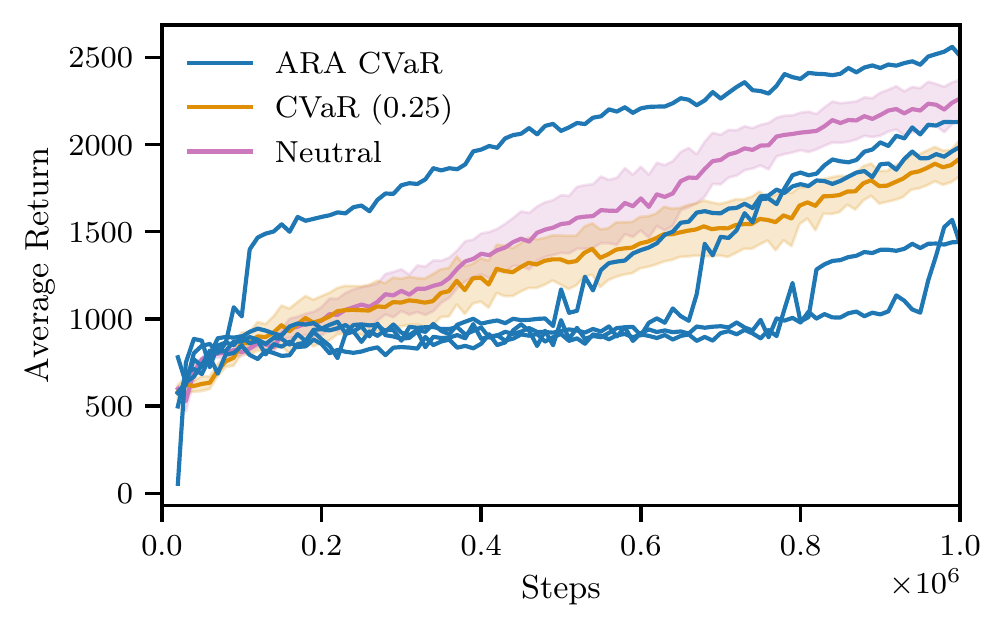}
    \includegraphics[width=0.45\linewidth]{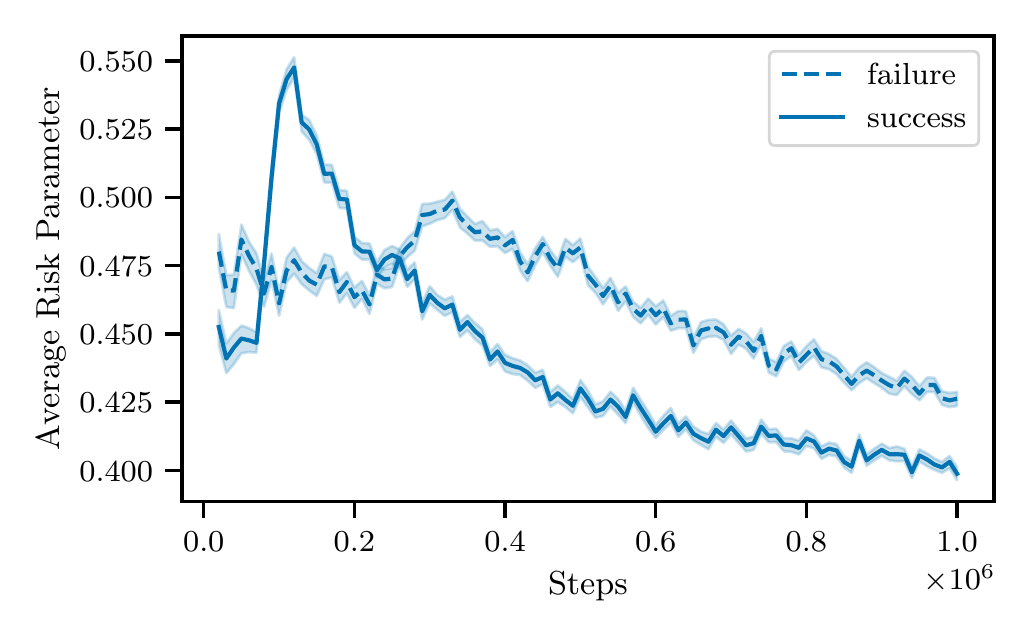}
\caption{Average risk parameter of 5 seeds of \ours{} during training in the \emph{AntBulletEnv-v0} environment \textbf{without} changing dynamics. 
There is a large performance gap between runs of \ours{} due to a lack of exploration. 
This is expressed by the difference in the risk level around step $150,000$ where the successful agent learns to walk, leading to new areas of the state space.
}
\label{fig:results_ant_risk}
\end{figure}

While we expect \ours{} to be an improvement over existing approaches in risk-aware DistRL, it is no silver bullet.
As previously noted by \cite{keramati-aaai20}, exploration heavy environments are hard to learn for risk-aware policies and we do not expect dynamic risk levels to solve this issue completely.
In fact, in \cref{fig:results_ant_risk} we can see the same exploration issue in \ours{} as in the conservative CVaR with $\alpha=0.25$ if we look at a single mass and friction setting of the Ant environment.
Runs that perform worse show a more conservative risk setting and are therefore impeding exploration.
Existing measures to mitigate the exploration problem should thus be considered in the future in combination with \ours{} if exploration is a limiting factor in a given environment.

Furthermore, although we remove the need to tune $\alpha$ itself, we also introduce a new hyperparameter by the RND architecture. 
It only depends on the overall distribution of observations, however, and is thus robust to the individual tasks that the agents learns to solve.
While we have found the original RND architecture to work well without any tuning for the settings presented in this paper, this may of course not be the case when applying \ours{} in new domains.


\section{Conclusion}
We showed that risk-aware quantile distortion policies with static $\alpha$ are suboptimal both in training and test settings across changing risk levels.
To solve this problem, we propose Automatic Risk Adaptation (\ours{}), which deploys Random Network Distillation (RND) as an uncertainty measure in order to automatically find the appropriate risk level for a given task.
We demonstrate that \ours{} can improve generalization and lower the failure rate of agents considerably compared to both risk-aware and non-risk-aware agents without adding a significant training overhead.
Our method makes no assumption about the specific distributional RL algorithm and is thus generally applicable to other methods such as Implicit Quantile Networks \citep{dabney-icml18}.
Tying risk-estimation to uncertainty is at the core of our approach and could be developed further in future work by e.g. developing more accurate representations of uncertainty or  exploring more complex mappings from uncertainty to risk-level that could make policies risk-aware only under certain conditions. 
As risk-aware RL is important to broad and reliable application of RL in the real-world, we believe these steps will contribute to improving the robustness of RL in the face of dynamic environments.



\newpage
\bibliographystyle{unsrtnat}
\bibliography{bib/strings, bib/lib, bib/rndiqn, bib/proc}

\section*{Checklist}


\begin{enumerate}

\item For all authors...
\begin{enumerate}
  \item Do the main claims made in the abstract and introduction accurately reflect the paper's contributions and scope?
    \answerYes{}
  \item Did you describe the limitations of your work?
    \answerYes{See Section \ref{limitations}}
  \item Did you discuss any potential negative societal impacts of your work?
    \answerYes{}
  \item Have you read the ethics review guidelines and ensured that your paper conforms to them?
    \answerYes{}
\end{enumerate}

\item If you are including theoretical results...
\begin{enumerate}
  \item Did you state the full set of assumptions of all theoretical results?
    \answerNA{}{}
	\item Did you include complete proofs of all theoretical results?
    \answerNA{}{}
\end{enumerate}

\item If you ran experiments...
\begin{enumerate}
  \item Did you include the code, data, and instructions needed to reproduce the main experimental results (either in the supplemental material or as a URL)?
    \answerYes{We include the code in the supplemental material}
  \item Did you specify all the training details (e.g., data splits, hyperparameters, how they were chosen)?
    \answerYes{Training details can be found in the Appendix \ref{app:hypers} as well as the code in the supplemental material}
	\item Did you report error bars (e.g., with respect to the random seed after running experiments multiple times)?
    \answerYes{We include standard deviation over 10 random seeds for our experiments}
	\item Did you include the total amount of compute and the type of resources used (e.g., type of GPUs, internal cluster, or cloud provider)?
    \answerYes{See Appendix \ref{app:hypers}}
\end{enumerate}

\item If you are using existing assets (e.g., code, data, models) or curating/releasing new assets...
\begin{enumerate}
  \item If your work uses existing assets, did you cite the creators?
    \answerYes{We use existing RL environments as well as extend an existing code base. We cite both,}
  \item Did you mention the license of the assets?
    \answerYes{}
  \item Did you include any new assets either in the supplemental material or as a URL?
    \answerNo{}
  \item Did you discuss whether and how consent was obtained from people whose data you're using/curating?
    \answerNA{}
  \item Did you discuss whether the data you are using/curating contains personally identifiable information or offensive content?
    \answerNA{}
\end{enumerate}

\item If you used crowdsourcing or conducted research with human subjects...
\begin{enumerate}
  \item Did you include the full text of instructions given to participants and screenshots, if applicable?
    \answerNA{}
  \item Did you describe any potential participant risks, with links to Institutional Review Board (IRB) approvals, if applicable?
    \answerNA{}
  \item Did you include the estimated hourly wage paid to participants and the total amount spent on participant compensation?
    \answerNA{}
\end{enumerate}

\end{enumerate}

\appendix

\section{Experiment Hardware \& Hyperparameters}
\label{app:hypers}

\paragraph{Hardware} All experiments with were conducted  on a slurm GPU cluster consisting of 6 nodes with eight Nvidia RTX 2080 Ti each. 
The maximum memory was 10GB. 

\paragraph{Hyperparameters} 

We used the same hyperparameters for all experiments. Following \cite{ma20}, we did not tune the policy temperature $\alpha$.

\begin{table}[h]
    \centering
    \begin{tabular}{l r}
        \toprule
        Hyperparameter & Value \\
        \midrule
        Policy Temperature & $0.2$ \\
        Learning Rate & $0.0003$ \\
        Batch Size & $256$ \\
        Discount Factor & $0.99$ \\
        Target Smoothing & $0.005$ \\
        Replay Buffer Size & $10^6$ \\
        Minimum Steps Before Training & $10^4$ \\
        Quantile Fractions & $32$ \\
        Quantile Fraction Embedding Size & $64$ \\
        Huber Regression Threshold & $1$ \\
        \bottomrule \\
    \end{tabular}
    \caption{Hyperparameters of DSAC}
    \label{tab:hyperparams_dsac}
\end{table}

The RND target architecture is the same as in \citep{burda-iclr19}, with $3$ convolutional layers followed by a dense layer with $512$ units, and the learning rate is set at $0.0003$. 
The predictor has the same architecture with two additional dense layers with $512$ units each.


\section{Extended Experiments on Non-Risky Environments}

\begin{figure}[h]
\centering
\begin{subfigure}[b]{0.49\linewidth}
    \centering
    \includegraphics[width=\linewidth]{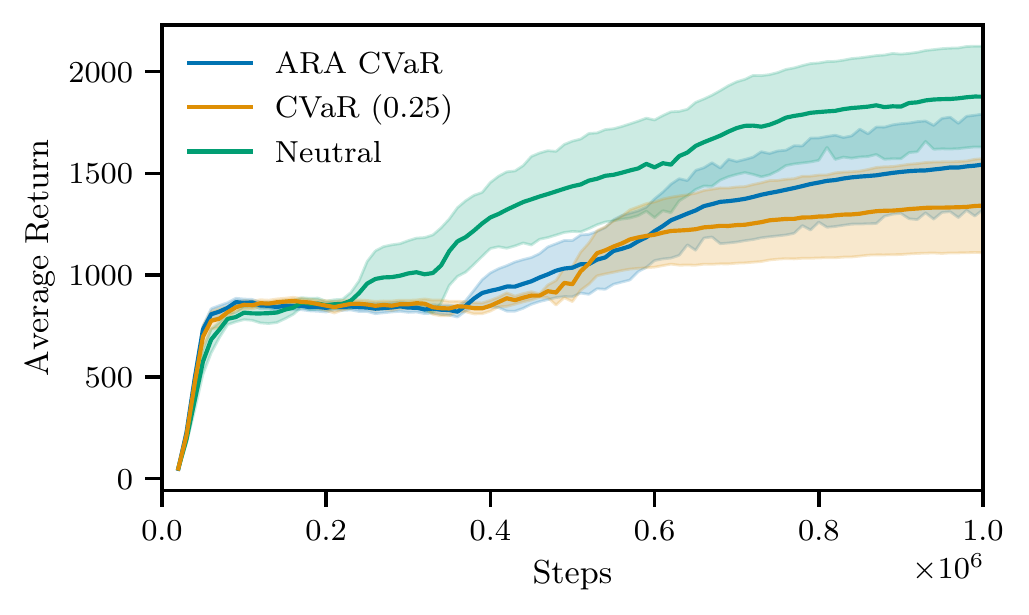}
\end{subfigure}
\begin{subfigure}[b]{0.49\linewidth}
    \centering
    \includegraphics[width=\linewidth]{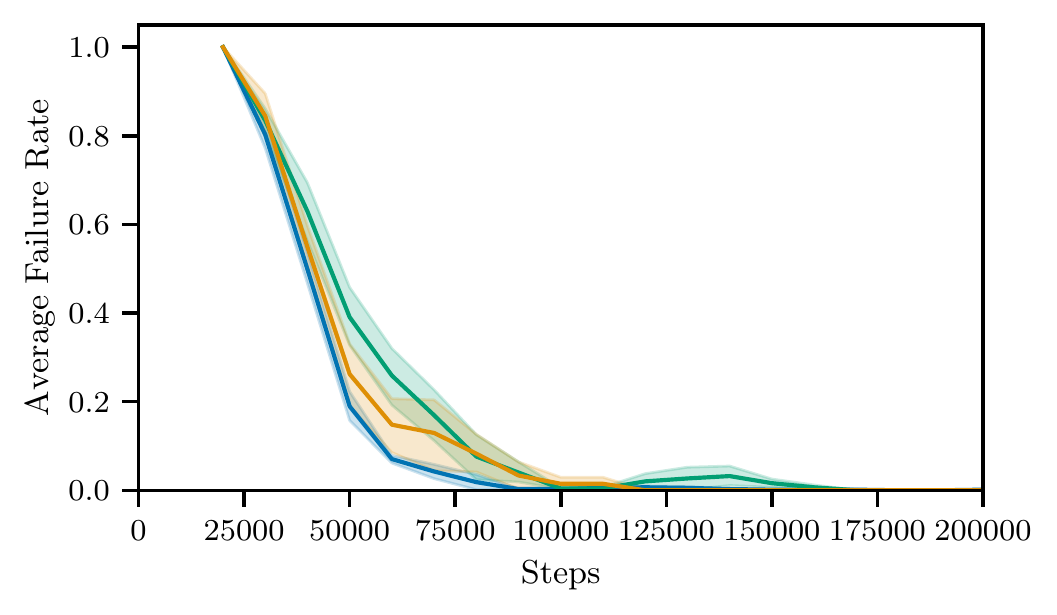}
\end{subfigure}
\caption{Average return and failure rate on \emph{Walker2DBulletEnv-v0} over 50 episodes (evaluated at each $10,000$ step interval) during training of Neutral DSAC, DSAC with fixed CVaR (0.25) and DSAC with \ours{}. Results are averaged over 5 seeds with the standard error indicated by the shaded area. We only show the failure rate over the first $200,000$ steps to better visualise the effect of \ours{}.}
\label{fig:results_walker2d}
\end{figure}

A task not mentioned so far with a risk of falling over is the \emph{Walker2DBulletEnv-v0} environment, which is based on the PyBullet physics engine \citep{pybullet}.
The result of \ours{} compared again with the two baselines is shown in \cref{fig:results_walker2d}.

As this environment is easier than \emph{BipedalWalkerHardcore-v3}, the agents learn to control the robot much faster and keep it from falling early on in training.
Here both risk-sensitive policies achieve low average failure rates faster than the neutral baseline, with \ours{} being ahead of the CVaR policy by about $1000$ steps.
Because \emph{Walker2DBulletEnv-v0} is more difficult in terms of exploration, their average return performance increases more slowly in turn.
However, as \ours{} adapts its risk level during training, it obtains higher rewards than the CVaR agent with a fixed risk level by about $23\%$, closing the gap to the neutral agent's performance by almost a third.

\section{Effect of the RND Mapping}
\label{app:rnd_mapping}

\begin{figure}[h]
    \centering
    \includegraphics[width=\textwidth]{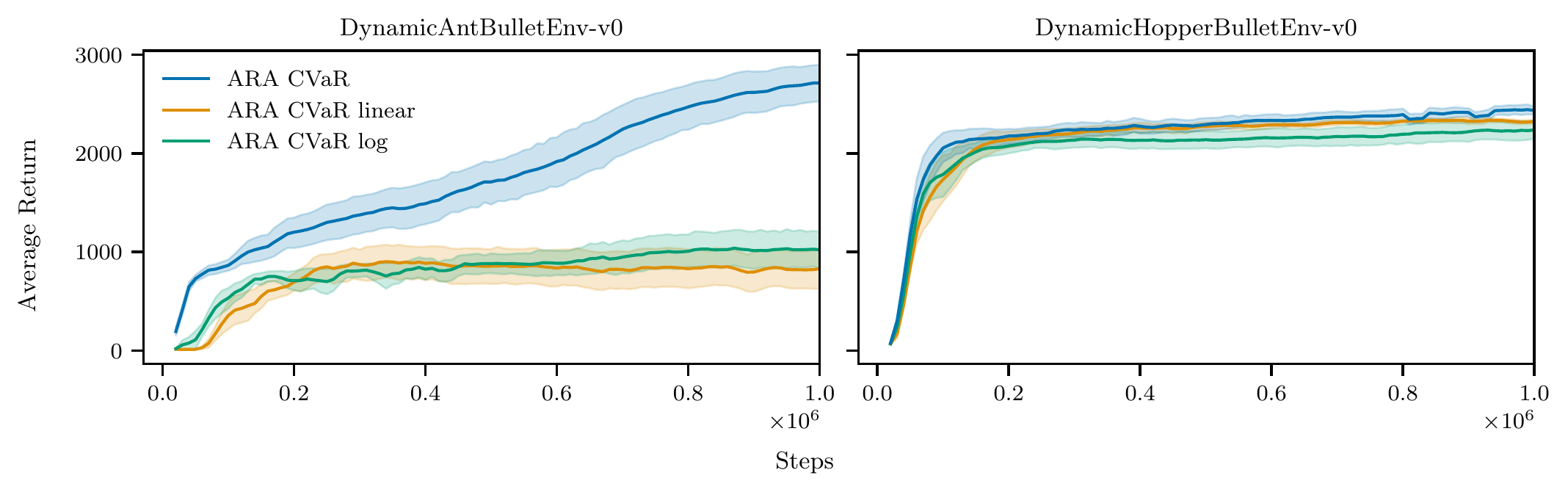}
    \caption{Average return over 50 episodes during training of \ours{} on the dynamic environments using different RND error mappings ($\pm$ standard error).}
    \label{fig:ablation_ant_hopper_log_linear}
\end{figure}

\begin{figure}[h]
    \centering
    \includegraphics[width=\textwidth]{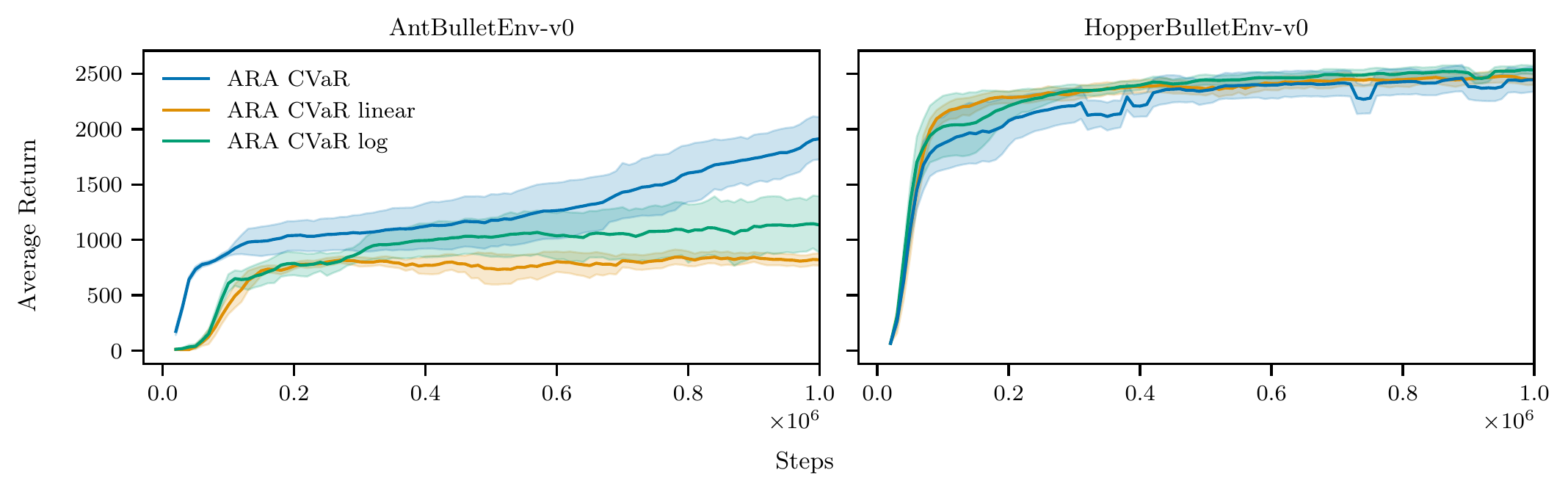}
    \caption{Average return over 50 episodes during training of \ours{} on the environments \textbf{without} changing dynamics using different RND error mappings ($\pm$ standard error).}
    \label{fig:ablation_ant_hopper_log_linear_nochange}
\end{figure}

We designed \ours{} with an exponential mapping from RND error to risk parameter as we believe it is important not to overlook possible failure states in safety-critical settings, even if we only see little uncertainty from the RND error. 
Comparing this original mapping with two alternatives, we can validate this intuition.

To contrast the original $\psi = e^{-u}$, we also evaluate a linear mapping in $\psi = -u + 1$ and a logarithmic one in $\psi = -u^2 + 1$.
While \ours{} is very sensitive even to small RND error values, the linear mapping behaves neutrally and the logarithmic mapping is less sensitive to small values for $u$, instead reacting more strongly for large ones.

We compare the three mappings on the \emph{AntBullet} and \emph{HopperBullet} environments, both with our $20\%$ friction and mass variations and without. 
As is the main paper, all results are averaged over $5$ random seeds.

In \cref{fig:ablation_ant_hopper_log_linear}  and \cref{fig:ablation_ant_hopper_log_linear_nochange} we can see that high sensitivity to uncertainty is beneficial both in the case of static and varying environment dynamics. 
While the performance on the \emph{HopperBullet} environment is very similar, the difference on the \emph{AntBullet} environment is considerable.

Both the linear and logarithmic versions hardly improve after the initial $200,000$ steps when training on changing environment dynamics, although logarithmic mapping is slightly better in the case of static dynamics.

In the case of the logarithmic mapping, this behaviour is to be expected. 
It leads to small uncertainties being largely ignored and generally acts in a more risk seeking way. 
We believe this is the reason for its slight performance increase on the static version of the \emph{AntBullet} environment as here the overall risk of falling is lower. 

Likely it is also the reason why the linear and logarithmic mappings perform so similar. 
The linear mapping places no particular importance on any value range of the RND error, which apparently results in a poor risk adaption for most cases.
The logarithmic mapping also cannot react appropriately to smaller uncertainties, but it seems better suited for states the agent knows well.

While an alternative logarithmic mapping might make sense in an explicitly risk seeking context, for our settings, the exponential version is by far the best choice.
As the linear mapping places no particular importance on either known or unknown states, it does not provide meaningful risk estimation for any situation and should therefore be avoided.

\section{Ablation}
\label{app:ablation}

\begin{figure}[h]
    \centering
    \includegraphics[width=\textwidth]{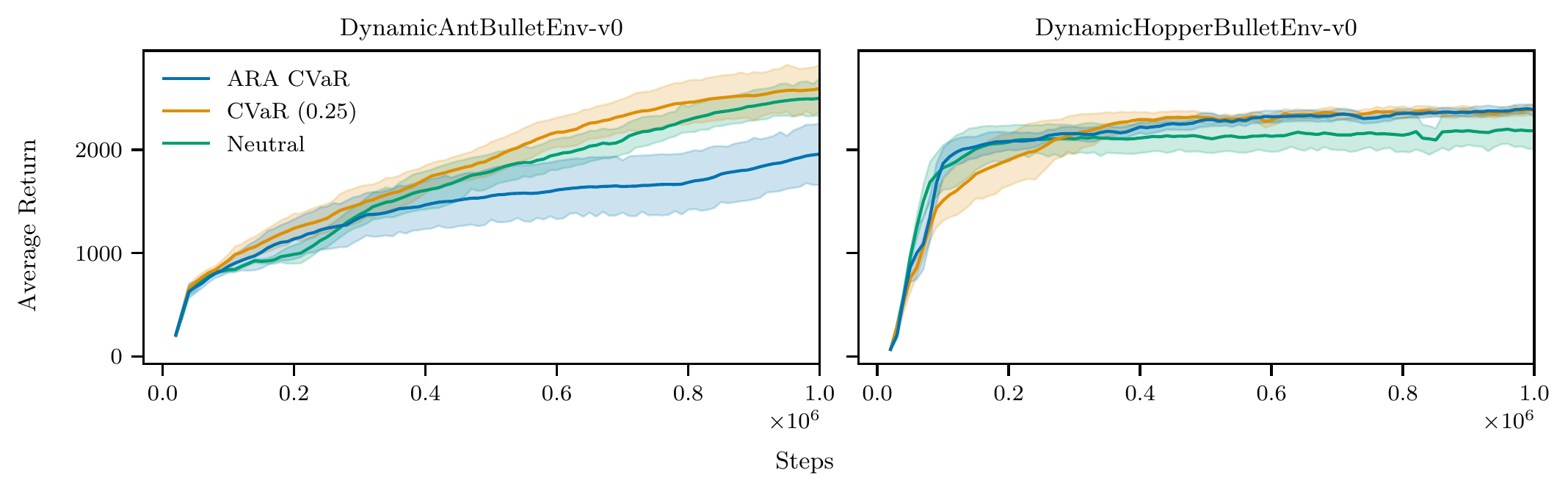}
    \caption{Average return over 50 episodes during training of \ours{} on dynamic environments with less variation ($\pm$ standard error).}
    \label{fig:ablation_nonrisky}
\end{figure}

Finally, we assessed how our choice of the strength of variation in the environment dynamics affects \ours{}.
In \cref{fig:ablation_nonrisky}, the friction and mass parameters are varied by only up to $\pm 10\%$.
In this setting, the inhibited exploration of \ours{} on the \emph{AntBullet} task is similar to the static environment which is expressed in the delayed increase in performance.

On \emph{HopperBullet}, \ours{} is able to achieve the same performance as CVaR with $\alpha$ of $0.25$ and is reaching an average return of $2000$ as fast as the neutral version.

These experiments indicate that \ours{} is more applicable in environments with a higher risk level but its automatic risk adaptation also makes it a viable solution in environments with less uncertainty.

\end{document}

%% file: rnd_macros.tex
\newcommand{\ours}{ARA}  
\newcommand{\pnorm}[1]{\left\lVert#1\right\rVert}

%% file: automl_macros.tex
\usepackage{amsmath}
\usepackage{mathtools}  
\usepackage{amsfonts}

\newcommand{\state}[0]{s}                               
\newcommand{\action}[0]{a}                              

